\documentclass[twoside,leqno,twocolumn]{article}

\usepackage[letterpaper]{geometry}

\usepackage{ltexpprt}
\usepackage{color}
\usepackage{amsmath,amssymb,amsfonts}
\usepackage{multirow}
\usepackage{array}
\usepackage{subfigure}
\usepackage{epsfig,epstopdf}
\usepackage{booktabs}
\usepackage{mathrsfs}
\usepackage{graphicx}
\usepackage{caption}
\usepackage{amsmath}
\usepackage{algorithm}
\usepackage{algorithmic}

\begin{document}

\title{\Large Meta Cross-Modal Hashing on Long-Tailed Data}
\author{Runmin Wang\thanks{College of Computer and Information Sciences, Southwest University, Chongqing, China.}
\and Guoxian Yu\thanks{School of Software, Shandong University. Corresponding Author: guoxian85@gmail.com}
\and Carlotta Domeniconi\thanks{Department of Computer Science, George Mason University, VA, USA.}
\and Xiangliang Zhang\thanks{CEMSE, King Abdullah University of Science and Technology, Thuwal, SA.}}

\date{}

\maketitle







\begin{abstract} \small\baselineskip=9pt Due to the advantage of reducing storage while speeding up query time on big heterogeneous data, cross-modal hashing has been extensively studied for approximate nearest neighbor search of multi-modal data. Most hashing methods  assume that training data is class-balanced.
However, in practice, real world data often have a long-tailed distribution.
In this paper, we introduce a meta-learning based cross-modal hashing method (MetaCMH) to handle long-tailed data. Due to the lack of training samples in the tail classes, MetaCMH first learns direct features from data in different modalities, and then introduces an associative memory module to learn the memory features of samples of the tail classes. It then combines the direct and memory features to obtain meta features for each sample. For samples of the head classes of the long tail distribution, the weight of the direct features is larger, because there are enough training data to learn them well; while for rare classes, the weight of the memory features is larger. Finally, MetaCMH uses a likelihood loss function to preserve the similarity in different modalities and learns hash functions in an end-to-end fashion. Experiments on long-tailed datasets show that MetaCMH performs significantly better than state-of-the-art methods, especially on the tail classes.
\end{abstract}

\section{Introduction}
With the rapid expansion of the Internet, a massive amount of multimedia data is being generated in our daily lives. To retrieve data in this scenario, approximate nearest neighbor (ANN) search plays an essential role in all kinds of applications. As a widely studied solution for ANN, hashing has sparked increasing research interest, due to its advantage in reducing storage consumption and speeding up query time \cite{Wang2016Learning, wang2018a}. Hashing aims at mapping  data vectors into fixed length binary codes, while preserving the proximity in the original space, and thus  facilitates approximate search in constant time.

Learning to hashing (LTH) is a popular category of hashing methods. LTH aims at learning an optimal compound hash function $\mathbf{b}=h(\mathbf{x})$ from the training data, and maps each input data sample onto a compact code vector $\mathbf{b}$. Therefore, LTH methods are data-dependent. Most hashing methods simply assume that the training data is balanced across different classes \cite{weiss2008spectral, gong2013iterative}. However, in the real world, the frequency distribution of visual classes is long-tailed \cite{reed2001the, liu2019large-scale}, which includes a few frequent (or head) classes and a large number of rare (or tail) ones. The  training samples of the tail classes are too scarce to learn the corresponding hashing function well. Thus, few-shot hashing methods have been proposed to deal with insufficient training samples \cite{gui2017few-shot, liu2019siamese-hashing}.

In many applications, the data has different modalities (e.g., image, text, video and audio) and we need to search for data items across the modalities. For example, we may want to find images or videos semantically related to the text of a query or to keywords. Thus, single-modal hashing methods are extended to multi-modal scenarios, and diverse cross-modal hashing methods have been proposed \cite{ding2014collective, xu2017attribute, liu2019cross, ji2020attribute-guided}. These methods neglect the problem of imbalanced classes and insufficient training samples in the training set. Cross-modal zero-shot learning methods have been proposed to model new classes, unseen in the training set but available in the test set \cite{ji2020attribute-guided, liu2019cross}; however, they still assume that the data distribution is balanced, or approximately balanced. In addition, these methods have been mainly applied to balanced datasets with only 10 to 30 different categories (e.g., \emph{Flickr}, one of the most widely used datasets, contains samples from 10 different categories). But in practice, the frequency distribution of visual classes is long-tailed \cite{reed2001the, liu2019large-scale}. Most of the existing cross-modal hashing methods only consider data samples in the head classes and ignore samples in the tail ones, which add up to a large portion of the whole distribution. Therefore, these methods show poor robustness on real world long-tailed data, as shown in our experiments.

Recently, meta-learning has been used extensively and has witnessed great success in a variety of fields \cite{vanschoren2018meta-learning:}. Meta-learning provides an alternative paradigm, where a machine learning method gains experience (meta knowledge) over learning episodes, to eventually boost its performance on few-shot data. In addition, it has been recognized that neural networks with memory capacities can empower meta-learning \cite{hochreiter2001learning}. Therefore, we use an associative memory module to solve the problem of few-shot learning and imbalanced classes simultaneously. Particularly, we embed the meta-learning paradigm into the cross-modal hashing framework, and propose a meta learning based cross-modal hashing method (MetaCMH). MetaCMH first introduces a memory-augmented feature learning network to transfer meta-knowledge from head classes with sufficient samples to tail ones with insufficient samples, and thus to learn meta features for samples from both head and tail classes. At the same time, MetaCMH minimizes a likelihood loss and a quantization loss to preserve the similarity in meta features and to learn hash codes of samples. In addition, a balance loss is used to make each bit of the hash code being balanced on all the training samples. Our main contributions are as follows:
\begin{itemize}
\item We consider the most common long-tailed data distribution and propose a novel CMH method (MetaCMH) to attack the problem of few-shot learning and imbalanced classes simultaneously.
\item We embed the meta-learning paradigm into the cross-modal hashing framework. We design an associative memory module to learn the meta-knowledge from samples of head classes and use this memory-based meta-knowledge to boost the performance on samples of tail classes.
\item The results on long-tailed datasets show that MetaCMH outperforms state-of-the-art cross-modal hashing methods, especially for the samples from tail classes.
\end{itemize}


\section{Related Works}
\label{relatedworks}
Our work is closely related to cross-modal hashing and meta-learning. We briefly review relevant recent work in these areas.
\subsection{Cross-modal hashing}
CMH methods can be divided into two main groups, supervised and unsupervised, based on whether they use the semantic (class) labels of the samples. Since in this paper we focus on the distribution of the number of samples from different classes, we also consider  zero-shot cross-modal hashing solutions, which can handle samples of new classes unseen during training.

Unsupervised CMH methods typically project the features of different modalities into a shared binary space to learn the hash functions. Unsupervised CMH was first inspired by canonical correlation analysis (CCA). For example, cross view hashing (CVH) \cite{kumar2011learning} aims to find a common Hamming space to maximize the correlation among different modalities. Collective matrix factorization hashing (CMFH) \cite{ding2014collective}
finds different projection functions from different modalities to a unified subspace, and takes these projection functions as hash functions.
Deep neural networks have achieved great success for their feature extraction capability. Thus, CMH methods with a deep architecture have attracted increasing interest in recent years. Unsupervised Deep Cross-Modal Hashing (UDCMH) \cite{wu2018unsupervised} integrates deep learning and matrix factorization with binary latent factor models, and explicitly constrains the hash codes to preserve the neighborhood structure of the original data. It obtains hash codes via solving a discrete-constrained objective function directly without relaxation.
The methods mentioned above are all unsupervised approaches which do not make use of valuable class labels of training data.

Supervised methods take advantage of semantic labels of training data and often achieve a  performance superior to unsupervised ones. To name a few, cross-modal similarity sensitive hashing (CMSSH) \cite{bronstein2010data} considers every bit of a hash code as a classification task, and learns the whole bits one by one. Semantic correlation maximization (SCM) \cite{zhang2014large-scale} optimizes the hashing functions by maximizing the correlation between two modalities with respect to the semantic similarity obtained from labels.
Semantics preserving hashing (SePH) \cite{lin2015semantics-preserving} transforms the semantic affinities obtained from class labels into a probability distribution and approximates this distribution with to-be-learned hash codes in the Hamming space, and finally learns the hash codes via kernel logistic regression.
Recently, supervised CMH methods combined with deep architectures have been extensively studied. For example, deep cross-modal hashing (DCMH) \cite{jiang2017deep} integrates feature learning and hash code learning into a same deep framework, and minimizes a joint loss function to preserve the similarity and generate hash codes simultaneously. Self-supervised adversarial hashing (SSAH) \cite{li2018self-supervised} introduces two adversarial networks and a uniform label-net to maximize the semantic correlation and consistency of the representations between different modalities.
Most of these CMH solutions assume that the labels are fully available during training, and the number of samples from different classes is balanced. In  real world scenarios, though, the labels of the training data are only partially known, or not fully reliable. For this reason, weakly-supervised CMH methods have been proposed. Weakly-supervised cross-modal Hashing (WCHash) \cite{8907427} enriches the labels of training data and defines a latent central modality to make efficient hashing on more than three modalities. Flexible cross-modal hashing (FlexCMH) \cite{2019Weakly} extends weakly-supervised cross-modal hashing on partially paired modality data by jointly optimizing the match between samples of different modalities and collaborative matrix factorization based hashing.

In real world applications, the number of samples from different classes is often imbalanced, and to expand cross-modal hashing to emerging new labels,  zero-shot CMH methods have been recently proposed. Attribute-guided network (AgNet) \cite{ji2020attribute-guided} aligns different modality data into a semantically rich attribute space via different attribute extraction networks, and then maps attribute vectors into hash codes. The shared attribute space enables the knowledge transfer from seen classes to unseen ones. Cross-modal zero-shot hashing (CZHash) \cite{liu2019cross} first learns a category space under the guidance of attribute data matrices, then learns hash functions via jointly optimizing the composite (inter-modal and intra-modal) similarity and category attribute space.

To the best of our knowledge, almost all the existing cross-modal hashing methods explicitly assume that the training set is class-balanced. As a result, they are vulnerable to long-tailed training sets, especially for samples from rare classes, as our experiments show. To overcome these limitations, we propose a novel CMH method (MetaCMH) based on meta learning for long-tail datasets.

\subsection{Meta-learning}
To date, no single or general definition of meta-learning exists \cite{vanschoren2018meta-learning:}. Intuitively, meta-learning is similar to transfer learning \cite{pan2010a}, where the model acquires experience (or meta-knowledge) over multiple learning episodes from source domains (data distributions) to improve the performance on new tasks with few training samples per class. Here we roughly group meta-learning methods into three categories according to the representation of meta-knowledge $\omega$. \\
\textbf{Parameter Initialization} here $\omega$ corresponds to the initial parameters of a neural network. For example,   model-agnostic meta-learning (MAML) \cite{maml2017} and probabilistic model-agnostic meta-learning (PMAML) \cite{finn2018probabilistic}
take meta-knowledge $\omega$ as initial conditions of the inner optimization. These methods aim to find a good initialization for the models, so that they can produce a good generalization performance on a new task with a small amount of training data. Latent embedding optimization (LEO) \cite{rusu2019meta} first learns a low-dimensional representation for the parameters and then does the meta-learning process in the low-dimensional subspace.\\
\textbf{Embedding Functions} first learn an embedding network $\omega$ in the meta-learning process and transform raw input data into a representation which can preserve the original data structure. Prototypical networks \cite{snell2017prototypical} learn a metric space in which few-shot classification is achieved by computing the Euclidean distances between prototype representations of each class.  Parameter networks \cite{qiao2018few-shot} assume that similar categories have similar parameters, and predict parameters from activations.\\
\textbf{Black-Box Models} directly train learner $\omega$ from the support set (i.e. $\theta=f_{\omega}(\mathcal{D}^{meta-train})$) rather than relying on the gradient. For example, memory-augmented neural networks can learn parameters from old data and assimilate new data quickly \cite{santoro2016meta-learning}. \cite{graves2014neural} changes the memory retrieval mechanism of Neural Turing Machines and introduces an external memory module to learn meta-knowledge in the training procedure. Then the meta-knowledge (memory parameter) is used to enhance the neural networks and improve the rapid generalization ability. Meta networks \cite{munkhdalai2017meta} improve on this neural model by combining slow weights and fast weights. Slow weights are learned by SGD like ordinary neural networks and the updates are slow. Fast weights are predicted by an extra neural network and improve the rapid generalization ability of the whole network.

Our MetaCMH builds on a memory-augmented network to learn meta features for each sample, so as to achieve the unification of samples from head classes and those from tail ones. In MetaCMH, the memory feature is not simply added as a direct feature with a learned weight, as in \cite{liu2019large-scale}. We determine the weights based on the extent to which the sample belongs to the tail class. This makes it more interpretable and  effective, as shown in the experiments.

\section{Proposed Method}
\label{method}
\subsection{Notation and Problem Definition}
Without loss of generality, in this paper we present the solution for two different modalities (i.e., image and text). Suppose that $\mathcal{O}=\{o_i\}^n_{i=1}$ is the dataset with $n$ samples, where each sample $o_i$ contains data from two different modalities $\{\mathbf{x}_i,\mathbf{y}_i\}$. Let $\mathbf{X}\in\mathbb{R}^{n\times d_x}$ denote the image modality and $\mathbf{x}_i\in \mathbb{R}^{d_x}$ corresponds to an image of $o_i$. Similarly, $\mathbf{Y}\in \mathbb{R}^{n\times d_y}$ is the text modality. For each sample $o_i$, we have a class label vector $\mathbf{l}_i\in\{0,1\}^L$ to encode which classes it belongs to, and $L$ is the number of distinct classes. If the $t$-th element of $\mathbf{l}_i$ is equal to 1, then $o_i$ belongs to class $t$; if it's equal to 0, $o_i$ does not belong to class $t$. Here, the number of samples in different classes subject to the long-tailed distribution (or power law distribution). In other words, a small number of classes has a moderate number of samples, and a large number of classes has a very small number of samples.
We denote the affinity (similarity) matrix as $\mathbf{A}$ to identify whether the samples are similar to one another. If sample $i$ and $j$ have at least one label in common, we assume they are similar to each other and set $a_{ij}=1$. On the other hand, if sample $i$ and sample $j$ do not have any label in common, we set $a_{ij}=0$.

The goal of cross-modal hashing is to find a series of hash functions for each modality to map samples into a fixed length binary code while preserving their original proximity. In this paper, we aim to find two hash functions $f_x(\mathbf{x})$ and $f_y(\mathbf{y})$ for the image and text modalities, where $f_x(\mathbf{x}_i)\in\{-1,+1\}^c$ is the hash code of sample $o_i$ obtained from the image modality, and $f_y(\mathbf{y}_i)\in\{-1,+1\}^c$ is the hash code obtained from the text modality. Here $c$ is the hash code length. With the cross-modal similarity preserved, if samples $o_i$ and $o_j$ are similar (i.e., $a_{ij}=1$), the Hamming distance between the hash codes $f_x(\mathbf{x}_i)$ and $f_y(\mathbf{y}_j)$ will be smaller; otherwise, the Hamming distance will be larger.

\begin{figure*} [h!tbp]
  \centering
  \includegraphics[width=15cm]{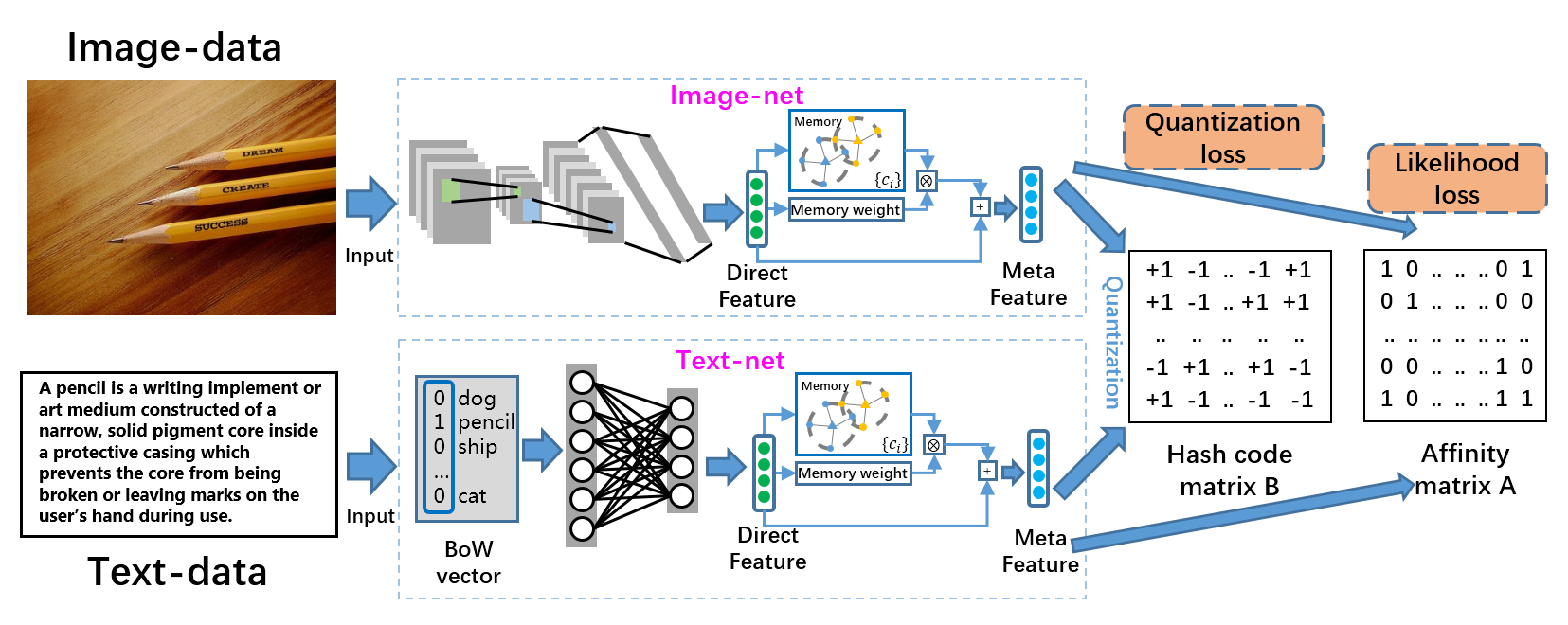}\\
  \caption{The MetaCMH framework. The meta embedding network for each modality consists of two parts, a basic network and a meta network. For each sample, we first obtain its direct features of the image (text) modality using a basic image (text) network, and use a memory-augmented meta network to enhance the direct features with meta features. We then quantify the meta features and generate hash codes for each sample.}
  \label{framework}
\end{figure*}

The whole framework of our method is illustrated in Figure \ref{framework}. There are two basic modules in our framework, the meta embedding and the hash code learning modules. The meta module first learns meta features for each sample from different modalities. The weights of memory features for samples from head classes will be small and that for samples from tail classes will be large, therefore we can enhance the feature extraction ability of our meta embedding module especially for tail classes, for which there are not sufficient training samples. We then obtain the hash code from our hash learning module, which preserves the similarity by minimizing a likelihood loss function and making hash codes of samples from the same class close to each other. The details of our meta network and hash code learning module are presented in the following subsections.

\subsection{Asymmetric Weighted Similarity}
\begin{figure} [!thb]
  \centering
  \includegraphics[width=7cm]{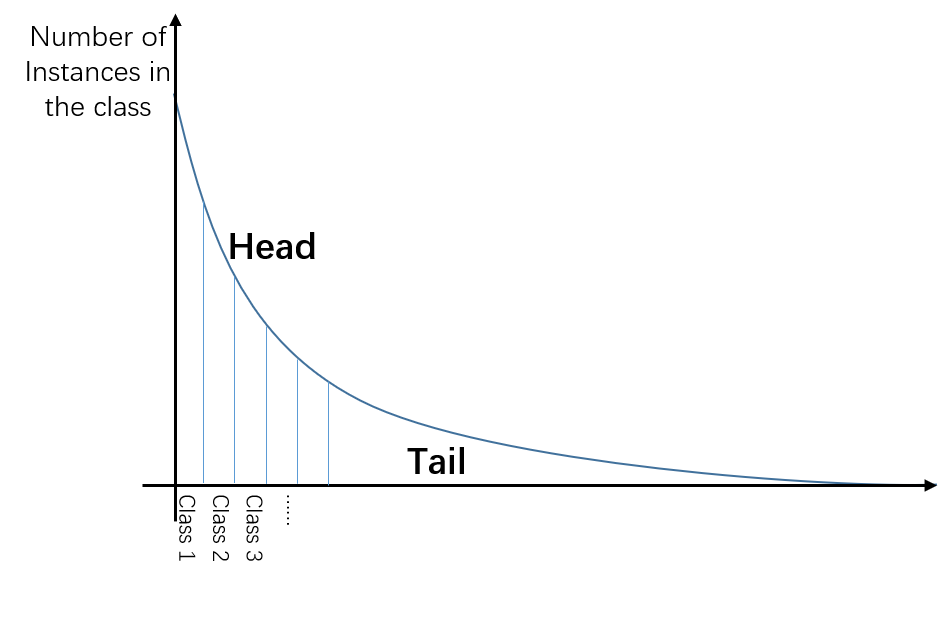}\\
  \caption{The distribution of instance numbers of different categories.}
  \label{f2}
\end{figure}

There are two main challenges in face of long-tailed data. As we can see in the Figure \ref{f2}, the first challenge is the category imbalance problem. Specifically, there are more instances in the head categories than tail categories, and this will lead to an imbalance weight of different categories in the model learning process. The second challenge is the few-shot learning problem in the tail categories which we will deal with in the later section.
To overcome the category imbalance problem, we propose an asymmetric weighted similarity.

\subsection{Meta Embedding}
In long-tailed data, samples can be roughly divided into two categories, samples from head classes and those from tail ones. Head classes have sufficient training samples, thus the neural network can learn their features well. For tail classes, the training samples are insufficient and ordinary neural networks perform poorly. Therefore, we use a meta embedding network to transfer meta-knowledge from head classes to tail ones, and enable our feature learning module to rapidly generalize to samples from tail ones. Firstly, we learn \emph{direct features} $\mathbf{v}_{direct}$ from each modality using a neural network. For the image modality, given the strong feature engineering ability of Convolutional Neural Networks, we choose a pre-trained CNN named CNN-F \cite{2014Return} to learn the features. CNN-F is a CNN with 8 layers. The detailed structure of the network is given in Table \ref{table1}. The first five layers are convolutional layers denoted by ``conv1-5''. In Table \ref{table1}, ``filters:$num\times size\times size$'' means there are $num$ convolution filters in the layer and each filter size is $size\times size$. ``st.$t$'' means the convolution stride is $t$. ``pad $n$'' means adding $n$ pixels to each edge of the input image. ``LRN'' means Local Response Normalisation is applied to this layer. ``pool'' denotes the down-sampling factor.
For text modality, we use a neural network with two full-connected layers. The detailed structure is also given in Table \ref{table1}. ``nodes:$n$'' means there are $n$ nodes in the layer. We want to remark that we mainly focus on the meta-embedding module, thus other alternative networks can be also adopted to extract features from image and text modalities.

CNN performs well under the prerequisite of sufficient training data \cite{krizhevsky2012imagenet}. But for long-tailed datasets, especially for samples from tail classes, we lack enough data for training. Given this, we introduce the \emph{memory features} $\mathbf{v}_{memory}$ to enhance the direct feature vector $\mathbf{v}_{direct}$. We know that the feature learning procedure projects the raw samples data into a semantic subspace. Samples of different classes may also have partially similar semantic attributes. For example, pandas and tigers belong to different classes but they are both furry. Thus, if we take the linear combination of prototypes of different classes as the memory features $\mathbf{v}_{memory}$, then the latter can be a good supplement for the direct features $\mathbf{v}_{direct}$. Inspired by this observation, we denote the meta  feature vector as $\mathbf{v}_{meta}$:
\begin{equation}
\small
\mathbf{v}_{meta} = \mathbf{v}_{direct} + \eta \mathbf{v}_{memory}
\label{vmeta}
\end{equation}
Similarly to LSTM \cite{hochreiter1997long}, in our meta-embedding network, both $\mathbf{v}_{memory}$ and $\mathbf{v}_{meta}$ are dependent on the input $\mathbf{v}_{direct}$.

Specifically, for the memory features $\mathbf{v}_{memory}$, we follow the idea in \cite{liu2019large-scale} and set $\mathbf{v}_{memory}$ to a linear combination of discriminative centroids (prototypes) of different classes.
We take the centroid of each class as an attribute and use their linear combination to enhance the direct features. We denote the centroid of the $i$-th class as $\mathbf{C}_i$, and $\mathbf{C}_i=\frac{1}{K}\sum_{j=1}^K \mathbf{v}_j$. Here $K$ is the number of samples of class $i$ and $\mathbf{v}_j$ is the direct feature vector for samples from class $i$. Then, we can define our memory feature vector as follows:
\begin{equation}
\small
\mathbf{v}_{memory}=\sum_i w_i \mathbf{C}_i
\label{vmemory}
\end{equation}
where $w_i$ is the weight of the prototype and it depends on $\mathbf{v}_{direct}$. Concretely, $w_i$ is learned by a shallow neural network whose input is $\mathbf{v}_{direct}$ and output is $w_i$. The parameters of this weight learning network $\theta_{\omega}$ are updated by SGD and back-propagation via minimizing the loss function, which is embedded in the learning process of the entire image-net or text-net.

In Eq. \eqref{vmeta}, $\eta$ is the weight for memory feature vector $\mathbf{v}_{memory}$. Intuitively, if the sample is from head classes, there are sufficient supervised training data in the same class, so the direct features $\mathbf{v}_{direct}$ can work well alone. Therefore, the weight of memory features $\eta$ should be smaller. On the other hand, if the sample is from tail classes, $\eta$ should be larger. Thus, we define the trade-off parameter as:
\begin{equation}
\small
\label{eq_eta}
\eta=\frac{d_{min}^{tail}}{d_{min}^{head}}
\end{equation}
where $d_{min}^{head} = \underset{j\in \mathcal{C}_{head}}{\min}|\mathbf{v}_i - \mathbf{C}_j|_2^2$ is the minimum distance to the prototypes of the head classes and $d_{min}^{tail} = \underset{j\in \mathcal{C}_{tail}}{\min}|\mathbf{v}_i - \mathbf{C}_j|_2^2$. In this way, we can ensure that the trade-off parameter $\eta_i$ for the samples of the head classes is small, while $\eta_i$ for the samples of the tail ones is large. In other meta-learning methods like \cite{liu2019large-scale}, a parameter similar to $\eta$ is learned by an extra shallow neural network. Our strategy is more interpretable and more effective, as shown in our experiments.

\begin{table}[]
\caption{Configuration of Image and Text networks}
\label{table1}
\centering
\small
\scalebox{0.9}{
\begin{tabular}{p{1.5cm}<{\centering}|p{0.8cm}<{\centering}|p{5cm}<{\centering}}
\hline
Network                    & Layers & Configuration                      \\\cline{1-3}
\multirow{11}{*}{Image-net} &\multirow{2}{*}{conv1} & filters:64 $\times$ 11 $\times$ 11, st. 4, pad 0, LRN, $\times$ 2 pool \\ \cline{2-3}
                       &\multirow{2}{*}{conv2}  & filters:256 $\times$ 5 $\times$ 5, st. 1, pad 2, LRN, $\times$ 2 pool   \\ \cline{2-3}
                       &{conv3}  & filters:256 $\times$ 3 $\times$ 3, st. 1, pad 1   \\ \cline{2-3}
                       &{conv4}  & filters:256 $\times$ 3 $\times$ 3, st. 1, pad 1   \\ \cline{2-3}
                       &\multirow{2}{*}{conv5}  & filters:256 $\times$ 3 $\times$ 3, st. 1, pad 1, $\times$ 2 pool  \\ \cline{2-3}
                           & fc6    & nodes:4096     \\ \cline{2-3}
                           & fc7    & nodes:4096     \\ \cline{2-3}
                           & fc8    & nodes:code length $c$             \\\cline{1-3}
\multirow{2}{*}{Text-net}  & fc1    & nodes:8192              \\ \cline{2-3}
                           & fc2    & nodes:code length $c$             \\\cline{1-3}
\end{tabular}}
\end{table}

\begin{figure} [!thb]
  \centering
  \includegraphics[width=7cm]{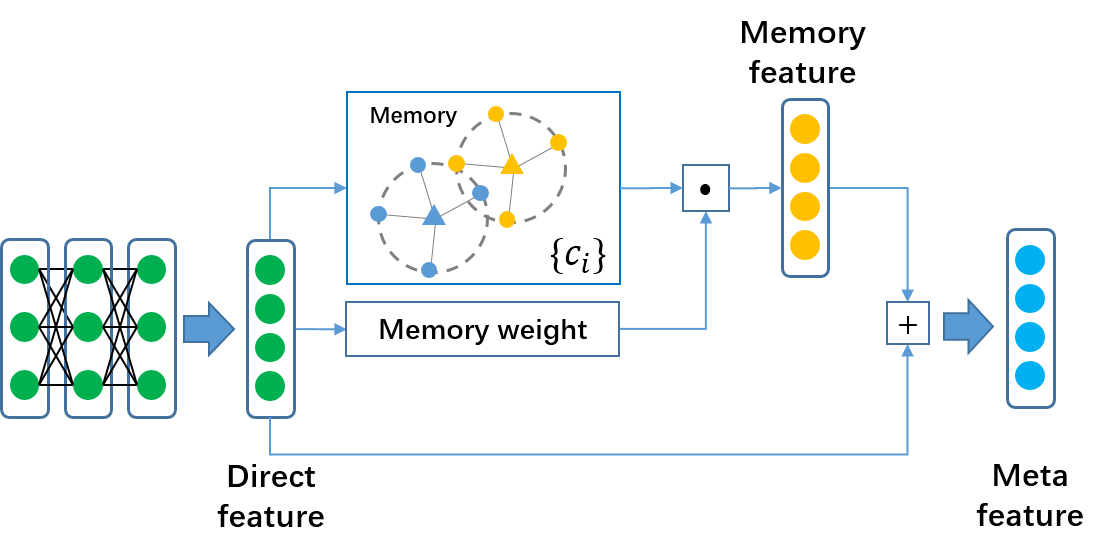}\\
  \caption{The meta feature learning network. We consider a linear combination of prototypes as memory features. The memory and direct features are combined to obtain the meta features.}
  \label{f1}
\end{figure}

\subsection{Hash code learning}
After extracting meta features for each sample, we then focus on how to obtain the hash code for each sample.
Suppose $\mathbf{A} \in \{0,1\}^{n\times n}$ is the affinity matrix of samples and each $a_{ij}=1$ in $\mathbf{A}$ means sample $i$ is similar to sample $j$, and $a_{ij}=0$ means sample $i$ is not similar to $j$. Here, we need to learn hash codes and hash functions while preserving the similarity information in $\mathbf{A}$. Suppose that $\mathbf{H}^{x}=[\mathbf{h}^x_1,\cdots,\mathbf{h}^x_n]\in\mathbb{R}^{c\times n}$ is the hash code matrix obtained from the image modality and $\mathbf{h}^x_i$ is the hash code obtained from the image modality for image $i$; $\mathbf{H}^{y}=[\mathbf{h}^y_1,\cdots,\mathbf{h}^y_n]\in\mathbb{R}^{c\times n}$ is the hash code matrix obtained from the text modality and $\mathbf{h}^y_i$ is the hash code obtained from the text modality for text $i$. Since $\mathbf{A}$ is known, we want to obtain the $\mathbf{H}^{x}$ and $\mathbf{H}^{x}$ corresponding to the highest probability. Therefore, we apply a maximum a posteriori approach:
\begin{equation}
\label{eq1}
\small
\begin{aligned}
\log p(\mathbf{H}^{x},\mathbf{H}^{y}|\mathbf{A}) &\propto \log p(\mathbf{A}|\mathbf{H}^{x},\mathbf{H}^{y})p(\mathbf{H}^{x})p(\mathbf{H}^{y}) \\
&=\sum_{i,j} \log p(a_{ij}|\mathbf{h}^{x}_{i},\mathbf{h}^{y}_{j})p(\mathbf{h}^{x}_{i})p(\mathbf{h}^{y}_{j})
\end{aligned}
\end{equation}
where $p(a_{ij}|\mathbf{h}^{x}_{i},\mathbf{h}^{y}_{j})$ is the likelihood function, $p(\mathbf{h}^{x})$ and $p(\mathbf{h}^{y})$ are prior probabilities of $\mathbf{h}^{x}$ and $\mathbf{h}^{y}$, which are unknown fixed values. Thus, we only consider the definition of the likelihood function $p(a_{ij}|\mathbf{h}^{x}_{i},\mathbf{h}^{y}_{j})$. As mentioned earlier, the principle of hashing is to preserve the proximity between samples, i.e., the smaller the Hamming distance between $\mathbf{h}^{x}_{i}$ and $\mathbf{h}^{y}_{j}$, the higher the probability that $a_{ij}$ will be 1.
Hamming distance is non-differentiable, thus we should find a surrogate. $\mathbf{h}^{x}_{i}$ and $\mathbf{h}^{y}_{j}$ here are both binary code vectors, so the Hamming distance between $\mathbf{h}^{x}_{i}$ and $\mathbf{h}^{y}_{j}$ can be converted into an inner product:
\begin{equation}
\small
dist_{H}(\mathbf{h}^{x}_{i},\mathbf{h}^{y}_{j})=\frac12(c-\left \langle \mathbf{h}^{x}_{i},\mathbf{h}^{y}_{j} \right \rangle)
\label{eq6}
\end{equation}
where $c$ is the length of hash codes, and $\left \langle \mathbf{h}^{x}_{i},\mathbf{h}^{y}_{j} \right \rangle$ denotes the inner product of $\mathbf{h}^{x}_{i}$ and $\mathbf{h}^{y}_{j}$. From Eq. (\ref{eq6}), we know that $dist_{H}(\mathbf{h}^{x}_{i},\mathbf{h}^{y}_{j})$ is negatively correlated with $\left \langle \mathbf{h}^{x}_{i},\mathbf{h}^{y}_{j} \right \rangle$. Thus, we define the likelihood function of $a_{ij}$ as follows:
\begin{equation}
\small
\label{eq2}
\begin{aligned}
p(a_{ij}|\mathbf{h}^{x}_{i},\mathbf{h}^{y}_{j})&=
\begin{cases}
\sigma(\phi_{ij}) &   a_{ij}=1   \\
1-\sigma(\phi_{ij}) &   a_{ij}=0
\end{cases}\\
&=\sigma(\phi_{ij})^{a_{ij}}(1-\sigma(\phi_{ij}))^{1-a_{ij}}
\end{aligned}
\end{equation}
where $\phi_{ij}=\frac{1}{2}\left \langle \mathbf{h}^{x}_{i},\mathbf{h}^{y}_{j} \right \rangle$ and $\sigma(\phi)=\frac{1}{1+e^{-\phi}}$ denotes the Sigmoid function. $\sigma(\phi)$ is positively correlated with argument $\phi$, so the smaller the Hamming distance between $\mathbf{h}^{x}_{i}$ and $\mathbf{h}^{y}_{j}$, the larger the inner product $\left \langle \mathbf{h}^{x}_{i},\mathbf{h}^{y}_{j} \right \rangle$ is. This results in a higher probability that $a_{ij}$ will be 1 and vice versa. Therefore, Eq. \eqref{eq2} is a good approximation of the likelihood function.

Let $\mathbf{v}^x_i=f^x(\mathbf{x}_i;\theta_x)\in\mathbb{R}^c$ and $\mathbf{v}^y_j=f^y(\mathbf{y}_j;\theta_y)\in\mathbb{R}^c$ denote the learned meta features of the image modality of sample $i$ and the text modality of sample $j$. $\theta_x$ and $\theta_y$ are the parameters of image-net and text-net, respectively. $\mathbf{V}^x=[\mathbf{v}^x_1,\cdots,\mathbf{v}^x_n]$ and $\mathbf{V}^y=[\mathbf{v}^y_1,\cdots,\mathbf{v}^y_n]$ are the feature matrices of the image modality and the text modality. We can define our loss function as follows:
\begin{equation}
\small
\begin{split}
\underset{\mathbf{B}, \theta_x, \theta_y}{min} \mathcal{J}=&-\sum_{i,j=1}^{n}(a_{ij}\Phi_{ij}-log(1+\mathrm{e}^{\Phi_{ij}}))\\
&+\alpha(||\mathbf{B}-\mathbf{V}^x||_F^2+||\mathbf{B}-\mathbf{V}^y||_F^2)\\
&+\beta(||\mathbf{V}^x\mathbf{1}||_F^2+||\mathbf{V}^y\mathbf{1}||_F^2)\\
s.t.\quad \mathbf{B} \in & \{+1, -1\}^{c\times n}
\end{split}
\label{lossfunction}
\end{equation}
where $\Phi_{ij}=\frac12\langle \mathbf{v}^x_i ,\mathbf{v}^y_j \rangle$ is the inner product between $\mathbf{v}^x_i$ and $\mathbf{v}^y_j$, $\mathbf{B}$ is the unified hash code. $\mathbf{1}$ is a vector with all elements being 1. The first term $-\sum_{i,j=1}^{n}(a_{ij}\Phi_{ij}-log(1+\mathrm{e}^{\Phi_{ij}}))$ is the negative log likelihood of the affinity matrix $\mathbf{A}$. It is easy to see that minimizing the first term can make the inner product between $\mathbf{v}_i^x$ and $\mathbf{v}_j^y$ small when $a_{ij}=0$, and large when $a_{ij}=1$. In other words, samples similar to each other will be assigned similar meta features. The second term $||\mathbf{B}-\mathbf{V}^x||_F^2+||\mathbf{B}-\mathbf{V}^y||_F^2$ is the quantization loss. Minimizing this term forces hash codes learnt from paired samples across modalities to be as similar as possible. The third term $||\mathbf{V}^x\mathbf{1}||_F^2+||\mathbf{V}^y\mathbf{1}||_F^2$ is the balance loss; minimizing this term can make the number of zeros and ones in the hash codes as close as possible, and thus enlarge the coding space.

\subsection{Optimization}
There are three parameters in our loss function Eq. \eqref{lossfunction}, $\mathbf{B}$, $\theta_x$, and $\theta_y$. It is hard to optimize these parameters simultaneously. Thus, we optimize these parameters by an alternative strategy. We iteratively learn one of the three parameters at a time while fixing the others and give the main idea here.\\
\textbf{Optimize $\mathbf{\theta_x}$($\mathbf{\theta_y}$)}. We learn the parameters of image-net $\theta_x$ with $\theta_y$ and keep $\mathbf{B}$ fixed. We first calculate the derivative of the loss function $\mathcal{J}$ with respect to the meta feature vector $\mathbf{v}_i^x$ for each sample $\mathbf{x}_i$, and then use the back-propagation (BP) algorithm and stochastic gradient descent (SGD) to update $\theta_x$ iteratively. $\frac{\partial\mathcal{J}}{\partial\mathbf{V}^x_{*i}}$ is calculated as follows:
\begin{equation}\label{eq7}
\small
\begin{split}
  \frac{\partial\mathcal{J}}{\partial\mathbf{V}^x_{*i}}=&\frac12\sum_{j=1}^{n}(\sigma(\Phi_{ij})\mathbf{V}^y_{*j}-a_{ij}\mathbf{V}^y_{*j})\\
  &+2\alpha(\mathbf{V}^x_{*i}-\mathbf{B}_{*i})+2\beta\mathbf{V}^x\mathbf{1}
\end{split}
\end{equation}
where $\mathbf{V}^x_{*i}$ denotes the $i$-th column of $\mathbf{V}^x$.
Then we can compute $\frac{\partial\mathcal{J}}{\partial\theta_x}$  using the chain rule, and then update $\theta_x$ based on BP.
The optimization of $\theta_y$ is similar.\\
\textbf{Optimize $\mathbf{B}$}. When $\theta_x$ and $\theta_y$ are fixed, the optimization problem can be reformulated as:
\begin{alignat}{1}
  \max_{\mathbf{B}}\quad & tr(\mathbf{B}(\mathbf{V}^x+\mathbf{V}^y)^T)\\
  \mbox{s.t.}\quad &\mathbf{B}\in\{+1,-1\}^{c\times n} \nonumber
\end{alignat}
We can easily solve this problem by letting $\mathbf{B}_{ij}$ have the same sign as $(\mathbf{V}^x+\mathbf{V}^y)_{ij}$:
\begin{equation}\label{B_cal}
\small
 \mathbf{B}=sign(\mathbf{V}^x+\mathbf{V}^y)
\end{equation}

\section{Experiments}
\label{experiments}
\subsection{Experimental setup}
We conduct our experiments on two different long-tailed datasets. As we stated earlier, most existing cross-modal hashing methods only focus on balanced data. There is no off-the-shelf benchmark multi-modal long-tailed dataset. Given this, we manually filter and modify the two datasets \emph{Flickr} \cite{huiskes2008the} and \emph{NUS-wide} \cite{chua2009nus-wide:}. The original \emph{Flickr} dataset contains of 25,015 images collected from the Flickr website. Each image is annotated with its corresponding textual tags and at least one label out of 24 different semantic labels. The textual tags of each image are represented with a 1386-dimensional BOW (bag of word) vector. Each image-tag pair is treated as an sample. For methods based on hand-crafted features rather than raw picture data, each image is converted to a 512-dimensional GIST vector. The original \emph{NUS-wide} dataset contains 269,648 images and corresponding textual tags collected from Flickr website. Each image-tag pair is treated as an sample and each sample is labeled with at least one label of 81 concepts. The textual tags are converted into a series of 1000-dimensional BOW vectors. For hand-crafted features based methods, each image is converted into a 500-dimensional bag-of-visual-words (BOVW) vector.

We convert these two datasets into long-tailed datasets in a similar way. Because most samples in the two datasets are labeled with not only one label, we first get rid of the excess labels on the samples. For samples with more than 3 labels, we randomly keep 2-3 high discriminative labels. Here we assume that class labels with fewer samples are more discriminative. For example, a tiger has two labels (``animal'' and ``feline''). Obviously,  more samples have the label ``animal'', and fewer the label ``feline''. To keep the tail labels, we select samples and construct our long-tailed training sets. We select a few classes with a large number of samples and a large number of classes with a few samples  (the specific quantities are shown in Table \ref{trainingset}). In this way, the number of samples in different classes follow a long-tailed distribution. As for the test (query) set, we select 50 samples from each class. We use the whole original dataset (excluding samples in the training set) as the retrieval set.

\begin{table}[t]
\caption{Distribution of head and tail classes of the training set.}
\label{trainingset}
\centering
\small
\begin{tabular}{c c c c c}
\hline
Dataset                    & Item                       &head  &tail1  &tail2                      \\\cline{1-5}
\multirow{2}{*}{Flickr}    & Number of classes          &4     &10    &10           \\ \cline{2-5}
                           & Samples in each class    &2000  &200   &50          \\\cline{1-5}
\multirow{2}{*}{NUS-wide}  & Number of classes          &9     &35    &35          \\ \cline{2-5}
                           & Samples in each class    &1000  &100   &25          \\\cline{1-5}
\end{tabular}
\end{table}

We use five representative methods for empirical comparison, including Deep Cross-Modal Hashing (DCMH) \cite{jiang2017deep}, Semantic Correlation Maximization (SCM-seq and SCM-orth) \cite{zhang2014large-scale}, Semantics Preserving Hashing (SePH) \cite{lin2015semantics}, and Collective Matrix Factorization Hashing (CMFH) \cite{ding2014collective}. SCM-seq, SCM-orth, and SePH are supervised methods with shallow architectures. DCMH and SDCH are supervised methods with deep architectures. CMFH is an unsupervised hash learning method. In addition, MetaCMH-v is introduced as a variant of MetaCMH and it uses a shallow neural network to learn the parameter $\eta$ rather than adopting Eq. \eqref{eq_eta}. For each comparison method, we use the hyper-parameters recommended in the corresponding article. For MetaCMH and MetaCMH-v, we set $\alpha=\beta=1$. We use the canonically used mean average precision (MAP) \cite{bronstein2010data, zhang2014large-scale} to quantify the performance of hashing retrieval methods.

\subsection{Results and Analysis}
We report the experimental results in Table \ref{result}, where ``I to T'' means the query is the image and the retrieval data is the text, and ``I to T'' means the query is the text and the retrieval data is the image. The standard deviations of MAP results of compared methods are quite small (generally less than 3\%), thus we do not report the standard deviations in the table.
From Table \ref{result}, we have the following observations and conclusions:

\begin{table*}[]
\caption{Mean Average Precision of all methods on Flickr and NUS-wide.}
\label{result}
\centering
\small
\scalebox{0.9}{
\begin{tabular}{c|l|llllll|llllll}
\hline
\multicolumn{2}{c|}{Dataset}        & \multicolumn{6}{c|}{Flickr}                                                                                                                                            & \multicolumn{6}{c}{NUS-wide}                                                                                                                                          \\ \hline
\multicolumn{2}{c|}{Class}          & \multicolumn{2}{c}{All classes}                           & \multicolumn{2}{c}{Head Classes}                              & \multicolumn{2}{c|}{Tail Classes}                               & \multicolumn{2}{c}{All classes}                           & \multicolumn{2}{c}{Head Classes}                              & \multicolumn{2}{c}{Tail Classes}                               \\ \hline
\multicolumn{2}{c|}{bits}           & \multicolumn{1}{c}{32bit} & \multicolumn{1}{c}{64bit} & \multicolumn{1}{c}{32bit} & \multicolumn{1}{c}{64bit} & \multicolumn{1}{c}{32bit} & \multicolumn{1}{c|}{64bit} & \multicolumn{1}{c}{32bit} & \multicolumn{1}{c}{64bit} & \multicolumn{1}{c}{32bit} & \multicolumn{1}{c}{64bit} & \multicolumn{1}{c}{32bit} & \multicolumn{1}{c}{64bit} \\ \hline
\multirow{7}{*}{I to T} & DCMH      & 0.492                     & 0.506                     & \textbf{0.597}                     & \textbf{0.601}                     & 0.351                     & 0.367                      & 0.387                     & 0.386                     & 0.496                     & 0.501                     & 0.258                     & 0.255                     \\
                        & CMFH      & 0.377                     & 0.382                     & 0.444                     & 0.446                     & 0.265                     & 0.274                      & 0.256                     & 0.261                     & 0.289                     & 0.302                     & 0.204                     & 0.215                     \\
                        & SCM-seq   & 0.387                     & 0.393                     & 0.451                     & 0.467                     & 0.255                     & 0.259                      & 0.264                     & 0.269                     & 0.304                     & 0.307                     & 0.213                     & 0.219                     \\
                        & SCM-orth  & 0.386                     & 0.392                     & 0.462                     & 0.477                     & 0.251                     & 0.255                      & 0.265                     & 0.271                     & 0.316                     & 0.319                     & 0.217                     & 0.225                     \\
                        & SePH      & 0.467                     & 0.475                     & 0.510                     & 0.521                     & 0.339                     & 0.346                      & 0.314                     & 0.326                     & 0.417                     & 0.421                     & 0.249                     & 0.254                     \\
                        & MetaCMH   & \textbf{0.521}                     & \textbf{0.527}                     & 0.581                     & 0.587                     & \textbf{0.485}                     & \textbf{0.491}                      & \textbf{0.431}                     & \textbf{0.440}                     & 0.505                     & 0.509                     & \textbf{0.307}                     & \textbf{0.319}                     \\
                        & MetaCMH-v & 0.501                     & 0.503                     & 0.590                     & 0.594                     & 0.401                     & 0.433                      & 0.420                     & 0.427                     & \textbf{0.507}                     & \textbf{0.514}                     & 0.287                     & 0.289                     \\ \hline
\multirow{7}{*}{T to I} & DCMH      & 0.510                     & 0.514                     & 0.583                     & 0.580                     & 0.376                     & 0.381                      & 0.413                     & 0.416                     & 0.541                     & 0.543                     & 0.303                     & 0.307                     \\
                        & CMFH      & 0.387                     & 0.396                     & 0.435                     & 0.437                     & 0.277                     & 0.284                      & 0.279                     & 0.287                     & 0.312                     & 0.322                     & 0.235                     & 0.239                     \\
                        & SCM-seq   & 0.399                     & 0.406                     & 0.491                     & 0.495                     & 0.261                     & 0.269                      & 0.295                     & 0.297                     & 0.327                     & 0.336                     & 0.247                     & 0.253                     \\
                        & SCM-orth  & 0.404                     & 0.413                     & 0.499                     & 0.501                     & 0.265                     & 0.276                      & 0.299                     & 0.312                     & 0.335                     & 0.336                     & 0.259                     & 0.264                     \\
                        & SePH      & 0.463                     & 0.475                     & 0.524                     & 0.526                     & 0.367                     & 0.374                      & 0.337                     & 0.346                     & 0.448                     & 0.457                     & 0.297                     & 0.305                     \\
                        & MetaCMH   & \textbf{0.541}                     & \textbf{0.543}                     & 0.588                     & 0.590                     & \textbf{0.495}                     & \textbf{0.503}                      & \textbf{0.458}                     & \textbf{0.462}                     & 0.538                     & 0.547                     & \textbf{0.356}                     & \textbf{0.359}                     \\
                        & MetaCMH-v & 0.522                     & 0.523                     & \textbf{0.591}                     & \textbf{0.593}                     & 0.404                     & 0.415                      & 0.435                     & 0.439                     & \textbf{0.541}                     & \textbf{0.550}                     & 0.290                     & 0.302                     \\ \hline
\end{tabular}}
\end{table*}

\noindent (i) The methods with a deep architecture (DCMH, MetaCMH and MetaCMH-v) outperform solutions with a shallow architecture (CMFH, SCM-seq, SCM-orth, SePH), and the former methods perform better than the latter ones on both head and tail classes. Therefore, methods with deep architectures achieve higher overall MAP values. This is mainly because the deep neural networks based methods have a stronger feature extraction capability than shallow ones.

\noindent (ii) Supervised methods perform better than unsupervised methods (CMFH). This is very intuitive, since supervised methods take advantage of class labels which can effectively improve the performance of hashing. The gap between CMFH and SCM-seq (or SCM-orth) is small because SCM simply uses a linear transformation to learn the hash function while trying to reconstruct the cosine similarity matrix with obtained hash codes, and it ignores structure information. In contrast, CMFH learns a latent semantic space via feature matrix decomposition, which can mine structure information.

\noindent (iii) MetaCMH uses a similar hash code learning method as DCMH, but DCMH only uses a basic deep neural network to learn the features of instances. We observe that, DCMH performs well on head classes, but it clearly loses to MetaCMH on tail ones. This is due to the fact that the memory-augmented network can transfer meta knowledge from head classes to tail ones, and can enhance the effectiveness on tail classes. We also observe that in some cases, the MAP values of DCMH on head classes are higher than those of MetaCMH. This is because for head classes with abundant training samples, the basic deep neural network can learn features of instances very well, and the extra memory features may in part damage the direct features.

\noindent (iv) MetaCMH-v achieves results comparable to MetaCMH on head classes, but it loses to MetaCMH on tail classes. This is because  MetaCMH-v uses a shallow neural network to learn the weight of memory features, which does not guarantee that the weights of memory features of instances from the head classes are significantly smaller than those from tail ones. As a result, direct features of instances from head classes are well learned, but those for tail classes are not, and the lower weights of instances from tail classes cannot well enhance the direct features.

Overall, these results and the ablation study confirm the necessity and effectiveness of MetaCMH on performing meta feature learning for long-tailed data.

\subsection{Parameter analysis}
\begin{figure}[htbp]
\centering
\subfigure[$\alpha$ vs. MAP]{
\begin{minipage}[t]{0.5\linewidth}
\centering
\includegraphics[width=4.5cm]{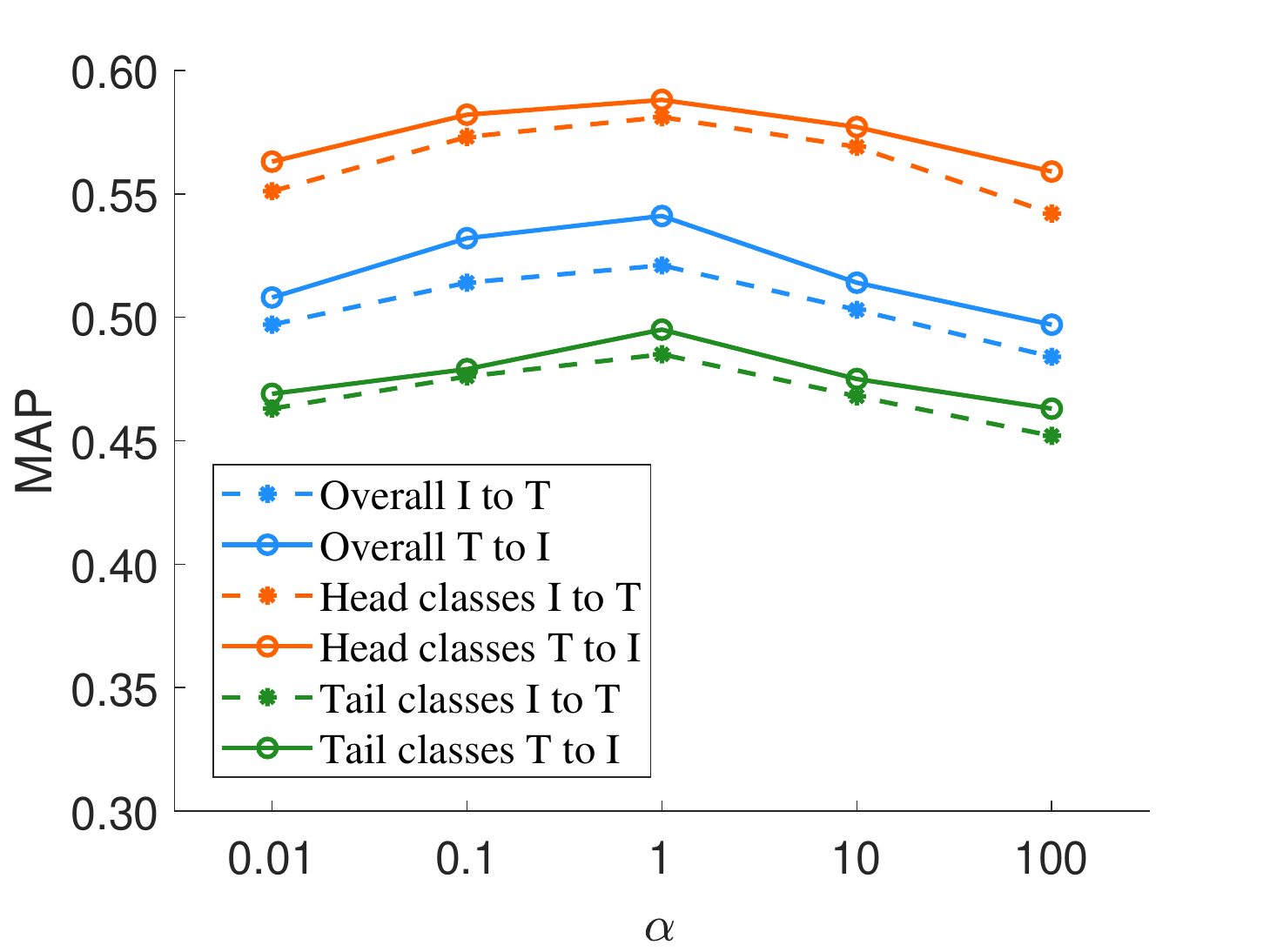}
\end{minipage}%
}%
\subfigure[$\beta$ vs. MAP]{
\begin{minipage}[t]{0.5\linewidth}
\centering
\includegraphics[width=4.5cm]{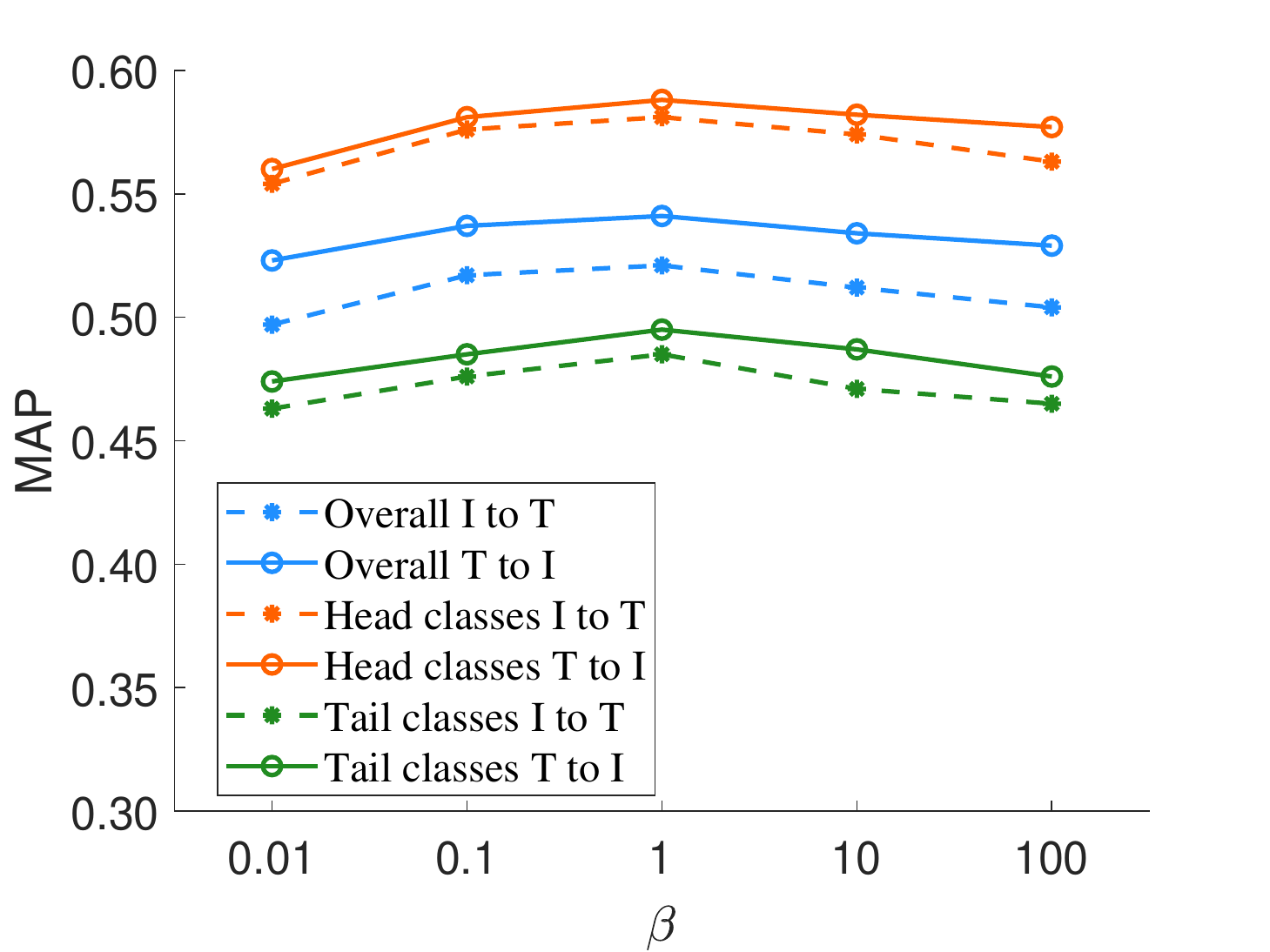}
\end{minipage}%
}%
\centering
\caption{Parameter sensitivity analysis of $\alpha$ and $\beta$ on the long-tailed \emph{Flickr} dataset.}
\label{parameter}
\end{figure}

In the MetaCMH model, there are two hyper-parameters: $\alpha$ and $\beta$. We conduct a series of experiments on the long-tailed \emph{Flicrk} dataset to study the influence of these two parameters. We set $\beta=1$ when testing the sensitivity of $\alpha$ and set $\alpha=1$ when testing the sensitivity of $\beta$. Results are shown in Figure \ref{parameter} where the hash code length is fixed to 32 bits.
We  observe that when $\alpha=1$, the MAP values reach their maximum values, and the model works best. As $\alpha$ increases or decreases, the effectiveness of the model decreases. Compared to $\alpha$, the model is less sensitive to the input value of $\beta$, but the overall pattern is similar as $\alpha$. When $\beta=1$, the MAP values reach their maximum values, and the model works best. Overall, a too small $\alpha$ cannot ensure  consistent hashing codes across modalities, while a too small $\beta$ cannot generate balanced binary codes. On the other hand, over-weighing them brings down the quality of cross-modal hashing functions.
From these observations, we adopt $\alpha=1$ and $\beta=1$ in the experiments.

\subsection{Ablation study}
In this section, we conduct a series of ablation study to validation the effectiveness of each components of the whole model.
Thus, we propose several variations of MetaCMH.

\section{Conclusion}
\label{conclusion}
In this paper, we proposed a novel cross-modal hashing method called Meta Cross-Modal Hashing (MetaCMH) for long-tailed data. MetaCMH introduces a meta-learning framework into cross-modal hashing and uses a memory-augmented network to transfer meta knowledge from head classes to tail ones. Our experiments on two long-tailed datasets show that meta-learning and CMH can work well together, and MetaCMH outperforms competitive methods on long-tailed datasets. The codes of MetaCMH are available at http://mlda.swu.edu.cn/codes.php?name=MetaCMH.

\bibliographystyle{unsrt}
\bibliography{meta-hashing}

\begin{thebibliography}{10}

\bibitem{Wang2016Learning}
Jun Wang, Wei Liu, Sanjiv Kumar, and Shih~Fu Chang.
\newblock Learning to hash for indexing big data - a survey.
\newblock {\em Proceedings of the IEEE}, 104(1):34--57, 2016.

\bibitem{wang2018a}
Jingdong Wang, Ting Zhang, Jingkuan Song, Nicu Sebe, and Heng~Tao Shen.
\newblock A survey on learning to hash.
\newblock {\em TPAMI}, 40(4):769--790, 2018.

\bibitem{weiss2008spectral}
Yair Weiss, Antonio Torralba, and Rob Fergus.
\newblock Spectral hashing.
\newblock In {\em NeurIPS}, pages 1753--1760, 2009.

\bibitem{gong2013iterative}
Yunchao Gong, Svetlana Lazebnik, Albert Gordo, and Florent Perronnin.
\newblock Iterative quantization: A procrustean approach to learning binary
  codes for large-scale image retrieval.
\newblock {\em TPAMI}, 35(12):2916--2929, 2013.

\bibitem{reed2001the}
William~J Reed.
\newblock The pareto, zipf and other power laws.
\newblock {\em Economics Letters}, 74(1):15--19, 2001.

\bibitem{liu2019large-scale}
Ziwei Liu, Zhongqi Miao, Xiaohang Zhan, Jiayun Wang, Boqing Gong, and Stella~X
  Yu.
\newblock Large-scale long-tailed recognition in an open world.
\newblock In {\em CVPR}, pages 2537--2546, 2019.

\bibitem{gui2017few-shot}
Yu-Xiong Wang, Liangke Gui, and Martial Hebert.
\newblock Few-shot hash learning for image retrieval.
\newblock In {\em ICCV}, pages 1228--1237, 2017.

\bibitem{liu2019siamese-hashing}
Chengcheng Liu, Huikai Shao, Dexing Zhong, and Jun Du.
\newblock Siamese-hashing network for few-shot palmprint recognition.
\newblock In {\em SSCI}, pages 3251--3258, 2019.

\bibitem{ding2014collective}
Guiguang Ding, Yuchen Guo, and Jile Zhou.
\newblock Collective matrix factorization hashing for multimodal data.
\newblock In {\em CVPR}, pages 2075--2082, 2014.

\bibitem{xu2017attribute}
Yahui Xu, Yang Yang, Fumin Shen, Xing Xu, Yuxuan Zhou, and Heng~Tao Shen.
\newblock Attribute hashing for zero-shot image retrieval.
\newblock In {\em ICME}, pages 133--138, 2017.

\bibitem{liu2019cross}
Xuanwu Liu, Zhao Li, Jun Wang, Guoxian Yu, Carlotta Domeniconi, and Xiangliang
  Zhang.
\newblock Cross-modal zero-shot hashing.
\newblock In {\em ICDM}, pages 449--458, 2019.

\bibitem{ji2020attribute-guided}
Z.~Ji, Y.~Sun, Y.~Yu, Y.~Pang, and J.~Han.
\newblock Attribute-guided network for cross-modal zero-shot hashing.
\newblock {\em IEEE Transactions on Neural Networks}, 31(1):321--330, 2020.

\bibitem{vanschoren2018meta-learning:}
Joaquin Vanschoren.
\newblock Meta-learning: A survey.
\newblock {\em arXiv preprint}, 2018.

\bibitem{hochreiter2001learning}
Sepp Hochreiter, A~Steven Younger, and Peter~R Conwell.
\newblock Learning to learn using gradient descent.
\newblock In {\em ICANN}, pages 87--94, 2001.

\bibitem{kumar2011learning}
Shaishav Kumar and Raghavendra Udupa.
\newblock Learning hash functions for cross-view similarity search.
\newblock In {\em IJCAI}, pages 1360--1365, 2011.

\bibitem{wu2018unsupervised}
G.~Wu, Z.~Lin, J.~Han, L.~Liu, G.~Ding, B.~Zhang, and J.~Shen.
\newblock Unsupervised deep hashing via binary latent factor models for
  large-scale cross-modal retrieval.
\newblock In {\em IJCAI}, pages 2854--2860, 2018.

\bibitem{bronstein2010data}
Michael~M Bronstein, Alexander~M Bronstein, Fabrice Michel, and Nikos Paragios.
\newblock Data fusion through cross-modality metric learning using
  similarity-sensitive hashing.
\newblock In {\em CVPR}, pages 3594--3601, 2010.

\bibitem{zhang2014large-scale}
Dongqing Zhang and Wujun Li.
\newblock Large-scale supervised multimodal hashing with semantic correlation
  maximization.
\newblock In {\em AAAI}, pages 2177--2183, 2014.

\bibitem{lin2015semantics-preserving}
Zijia Lin, Guiguang Ding, Mingqing Hu, and Jianmin Wang.
\newblock Semantics-preserving hashing for cross-view retrieval.
\newblock In {\em CVPR}, pages 3864--3872, 2015.

\bibitem{jiang2017deep}
Qing{-}Yuan Jiang and Wu{-}Jun Li.
\newblock Deep cross-modal hashing.
\newblock In {\em CVPR}, pages 3232--3240, 2017.

\bibitem{li2018self-supervised}
C.~Li, C.~Deng, N.~Li, W.~Liu, X.~Gao, and D.~Tao.
\newblock Self-supervised adversarial hashing networks for cross-modal
  retrieval.
\newblock In {\em CVPR}, pages 4242--4251, 2018.

\bibitem{8907427}
X.~{Liu}, G.~{Yu}, C.~{Domeniconi}, J.~{Wang}, G.~{Xiao}, and M.~{Guo}.
\newblock Weakly-supervised cross-modal hashing.
\newblock {\em TBD}, 99(1):1--14, 2020.

\bibitem{2019Weakly}
Xuanwu Liu, Jun Wang, Guoxian Yu, Carlotta Domeniconi, and Xiangliang Zhang.
\newblock Weakly-paired cross-modal hashing.
\newblock {\em TNNLS}, 99(1):1--11, 2020.

\bibitem{pan2010a}
Sinno~Jialin Pan and Qiang Yang.
\newblock A survey on transfer learning.
\newblock {\em TKDE}, 22(10):1345--1359, 2010.

\bibitem{maml2017}
Chelsea Finn, Pieter Abbeel, and Sergey Levine.
\newblock Model-agnostic meta-learning for fast adaptation of deep networks.
\newblock In {\em ICML}, page 1126¨C1135, 2017.

\bibitem{finn2018probabilistic}
Chelsea Finn, Kelvin Xu, and Sergey Levine.
\newblock Probabilistic model-agnostic meta-learning.
\newblock In {\em NeurIPS}, pages 9516--9527, 2018.

\bibitem{rusu2019meta}
Andrei~A Rusu, Dushyant Rao, Jakub Sygnowski, Oriol Vinyals, Razvan Pascanu,
  Simon Osindero, and Raia Hadsell.
\newblock Meta-learning with latent embedding optimization.
\newblock In {\em ICRL}, 2019.

\bibitem{snell2017prototypical}
Jake Snell, Kevin Swersky, and Richard~S Zemel.
\newblock Prototypical networks for few-shot learning.
\newblock In {\em NIPS}, pages 4077--4087, 2017.

\bibitem{qiao2018few-shot}
Siyuan Qiao, Chenxi Liu, Wei Shen, and Alan~L Yuille.
\newblock Few-shot image recognition by predicting parameters from activations.
\newblock In {\em CVPR}, pages 7229--7238, 2018.

\bibitem{santoro2016meta-learning}
Adam Santoro, Sergey Bartunov, Matthew Botvinick, Daan Wierstra, and Timothy
  Lillicrap.
\newblock Meta-learning with memory-augmented neural networks.
\newblock In {\em ICML}, pages 1842--1850, 2016.

\bibitem{graves2014neural}
Alex Graves, Greg Wayne, and Ivo Danihelka.
\newblock Neural turing machines.
\newblock {\em arXiv: Neural and Evolutionary Computing}, 2014.

\bibitem{munkhdalai2017meta}
Tsendsuren Munkhdalai and Hong Yu.
\newblock Meta networks.
\newblock In {\em ICML}, pages 2554--2563, 2017.

\bibitem{2014Return}
Ken Chatfield, Karen Simonyan, Andrea Vedaldi, and Andrew Zisserman.
\newblock Return of the devil in the details: Delving deep into convolutional
  nets.
\newblock In {\em British Machine Vision Conference}, pages 1--12, 2014.

\bibitem{krizhevsky2012imagenet}
Alex Krizhevsky, Ilya Sutskever, and Geoffrey~E Hinton.
\newblock Imagenet classification with deep convolutional neural networks.
\newblock In {\em NeurIPS}, pages 1097--1105, 2012.

\bibitem{hochreiter1997long}
Sepp Hochreiter and Jurgen Schmidhuber.
\newblock Long short-term memory.
\newblock {\em Neural Computation}, 9(8):1735--1780, 1997.

\bibitem{huiskes2008the}
Mark~J Huiskes and Michael~S Lew.
\newblock The mir flickr retrieval evaluation.
\newblock In {\em ACM MIRM}, pages 39--43, 2008.

\bibitem{chua2009nus-wide:}
Tatseng Chua, Jinhui Tang, Richang Hong, Haojie Li, Zhiping Luo, and Yantao
  Zheng.
\newblock Nus-wide: a real-world web image database from national university of
  singapore.
\newblock In {\em CIVR}, pages 48--52, 2009.

\bibitem{lin2015semantics}
Zijia Lin, Guiguang Ding, Mingqing Hu, and Jianmin Wang.
\newblock Semantics-preserving hashing for cross-view retrieval.
\newblock In {\em CVPR}, pages 3864--3872, 2015.

\end{thebibliography}
\end{document}